\documentclass[10pt,letterpaper]{article}

\usepackage[margin=1in]{geometry}  
\usepackage{mathptmx}              

\usepackage{booktabs}              
\usepackage{tabularx}              
\usepackage[table]{xcolor}         
\definecolor{lightgreen}{RGB}{220,255,220}
\definecolor{lightred}{RGB}{255,220,220}

\usepackage{graphicx}              
\usepackage{caption}               
\usepackage[numbers]{natbib}  

\usepackage{enumitem}              

\usepackage[colorlinks=true,linkcolor=blue,citecolor=blue,urlcolor=blue]{hyperref}

\usepackage{amsmath,amssymb}       

\newcolumntype{L}[1]{>{\raggedright\arraybackslash}p{#1}}

\usepackage{float}                 

\usepackage{adjustbox}

\definecolor{lightgreen}{RGB}{220,255,220}
\definecolor{lightred}{RGB}{255,220,220}

\begin{document}

\title{Beyond ImageNet: Understanding Cross-Dataset Robustness of Lightweight Vision Models}

\author{
  Weidong Zhang\thanks{Email: weidong.zhang@savvas.com} \\
  Department of Computer Science \\
  Arizona State University
  \and
  Pak Lun Kevin Ding\thanks{Email: kevinding@asu.edu} \\
  Department of Computer Science \\
  Arizona State University
  \and
  Huan Liu\thanks{Email: huanliu@asu.edu} \\
  Department of Computer Science \\
  Arizona State University
}

\date{}
\maketitle

\begin{abstract}

Lightweight vision classification models such as MobileNet, ShuffleNet, and EfficientNet are increasingly deployed in mobile and embedded systems, yet their performance has been predominantly benchmarked on ImageNet. This raises critical questions: Do models that excel on ImageNet also generalize across other domains? How can cross-dataset robustness be systematically quantified? And which architectural elements consistently drive generalization under tight resource constraints? Here, we present the first systematic evaluation of 11 lightweight vision models ($\sim$2.5M parameters), trained under a fixed 100-epoch schedule across 7 diverse datasets. We introduce the Cross-Dataset Score ($xScore$), a unified metric that quantifies the consistency and robustness of model performance across diverse visual domains. Our results show that (1) ImageNet accuracy does not reliably predict performance on fine-grained or medical datasets, (2) xScore provides a scalable predictor of mobile model performance that can be estimated from just four datasets, and (3) certain architectural components—such as isotropic convolutions with higher spatial resolution and channel-wise attention—promote broader generalization, while Transformer-based blocks yield little additional benefit, despite incurring higher parameter overhead. This study provides a reproducible framework for evaluating lightweight vision models beyond ImageNet, highlights key design principles for mobile-friendly architectures, and guides the development of future models that generalize robustly across diverse application domains.

\end{abstract}

\noindent\textbf{Index Terms—}Lightweight vision models, mobile AI, edge AI, cross-dataset generalization, deep learning architectures, performance benchmarking

\section{Introduction}

\subsection{Motivation}

The proliferation of mobile and edge devices has created an urgent demand for lightweight yet high-performing deep learning models capable of operating under strict computational, memory, and energy constraints. In response, the research community has developed a variety of mobile-oriented architectures for vision classification tasks~\cite{han2020ghostnet, howard2017mobilenets,mehta2021mobilevit,tan2019efficientnet, vasu2023mobileone, zhang2018shufflenet}, along with techniques such as pruning, quantization, and neural architecture search (NAS)~\cite{liu2024lightweight, menghani2023efficient,wu2019fbnet,zhou2020rethinking} to further optimize efficiency for resource-constrained environments. These advances have significantly improved model compactness and computational efficiency, enabling real-time AI capabilities on smartphones, IoT devices, and embedded systems where low latency and energy efficiency are critical~\cite{wang2025optimizingedgeai,wang2025empowering,lin2023tinyml}.

Mobile vision classification serves as a core capability for a wide range of applications—from camera-based recognition and on-device scene understanding to visual assistance and industrial inspection systems—making it a fundamental benchmark for edge intelligence research~\cite{fang2023progressimagenet,tuggener2022enough,vishniakov2023convnet}. Despite rapid advances in efficient architecture design and compression techniques, much of the optimization remains tightly coupled to ImageNet-scale benchmarks, raising questions about how well these architectures generalize to unseen domains~\cite{kornblith2019transfer}. This gap motivates a systematic investigation into the cross-dataset generalization of mobile-friendly vision models under standardized training conditions, without task-specific fine-tuning or additional parameters.

\subsection{Problem}

Despite being specifically designed for resource-limited environments and often targeting simpler vision classification tasks, mobile models are predominantly benchmarked on \textbf{ImageNet}---the de facto standard for measuring accuracy and efficiency~\cite{deng2009imagenet,krizhevsky2012imagenet}. Achieving competitive ImageNet performance, however, often demands models with larger parameter counts and deeper hierarchies, directly contradicting the core design philosophy of compact, deployable architectures~\cite{han2020ghostnet, howard2017mobilenets,zhang2018shufflenet}. This paradox exposes several key research questions:

\begin{itemize}
        
\item \textbf{Do Mobile Architectures Generalize or Mainly Specialize?}
When trained from scratch under identical and limited parameter budgets and optimization constraints, can a mobile architecture maintain consistent performance across diverse datasets and visual domains? Or do dataset-specific tuning and NAS-driven optimizations induce overfitting, limiting its ability to generalize beyond the dataset it was designed for?

\item \textbf{Is ImageNet a Fair Benchmark for Mobile Models?}
Are mobile architectures fundamentally suited for training on large-scale datasets such as ImageNet, or do smaller yet diverse datasets provide a more realistic and efficient foundation for benchmarking and evaluation? More broadly, how can we establish a fair and scalable metric that balances accuracy, robustness, and computational efficiency under constrained resources?   

\item \textbf{What Defines Robustness in Mobile Vision Architectures?}
Which architectural design principles—such as channel expansion, attention mechanisms, or spatial resolution preservation—consistently promote cross-domain robustness within tight parameter and FLOP budgets? And how can these insights guide the next generation of lightweight, general-purpose mobile vision models?

\end{itemize}

\subsection{Approach}

To address these questions, we perform a systematic evaluation of 11 representative mobile-friendly vision architectures across 7 diverse datasets, training all models from scratch under a unified experimental protocol. This design decouples the influence of architectural choices from confounding factors such as data preprocessing, training schedules, optimizer and loss function selection. Our analysis is structured around three complementary objectives:

\begin{itemize}
\item \textbf{Architecture suitability and training scalability:} 
We examine whether mobile architectures sustain consistent performance when trained on datasets of varying scale and complexity, and assess whether compact datasets can serve as reliable surrogates for large-scale ImageNet training~\cite{deng2009imagenet}—a key consideration given the computational demands of full ImageNet optimization.
    
\item \textbf{Performance extensibility and robustness:} We quantify  how well ImageNet-optimized architectures generalize to unseen domains under standardized training settings—without model-specific fine-tuning or additional parameters~\cite{fang2023progressimagenet,kornblith2019transfer}.

\item \textbf{Fair evaluation and design guidance:} We establish a level playing field by enforcing consistent parameter budgets, input resolutions ($224\times224$), and augmentation policies (e.g., CutMix~\cite{yun2019cutmix}). This enables fair cross-architecture comparisons and facilitates identification of design elements that promote robust generalization under resource constraints~\cite{fang2023progressimagenet,vishniakov2023convnet}.
\end{itemize}

\subsection{Contributions}
Our work makes the following key contributions:  

\begin{itemize}
    \item \textbf{Unified benchmark and framework.}  
       We establish a reproducible benchmark for mobile-friendly vision research by evaluating 11 representative models across 7 diverse datasets under an identical training protocol. This controlled setup disentangles architectural advantages from training-related confounders, enabling fair comparisons across CNN, NAS-based, and hybrid Transformer families. All code, training configurations, and evaluation scripts are released as an extensible framework to facilitate future research~\cite{beyond_imagenet_repo}.
    
    \item \textbf{Cross-dataset robustness and evaluation metric.}  
    We introduce the \textbf{xScore} metric, a principled measure of cross-dataset generalization that captures both accuracy and consistency, highlighting when ImageNet-derived gains translate effectively to other domains. This framework enables systematic assessment of performance extensibility and robustness across diverse datasets.
    
    \item \textbf{Architectural insights and design guidance.}  
    We identify key architectural elements that correlate with strong generalization under resource constraints, providing actionable guidance for designing the next generation of efficient, mobile-friendly models.      
\end{itemize}

\section{Related Work}

\textsc{ImageNet and Transferability.} Several studies question ImageNet's adequacy as a predictor of real-world performance. Kornblith et al.~\cite{kornblith2019transfer} show that higher ImageNet accuracy does not always translate to proportional gains on downstream datasets, highlighting its limitations as a proxy for generalization. Similarly, Fang et al.~\cite{fang2023progressimagenet} report that improvements on ImageNet do not reliably transfer to specialized domains such as medical imaging~\cite{tschandl2018ham10000} or scene recognition~\cite{xiao2010sun, quattoni2009recognizing}. Vishniakov et al.~\cite{vishniakov2023convnet} further demonstrate that models with similar ImageNet performance, including ConvNeXt and ViT under both supervised and CLIP pretraining, can diverge markedly in transferability, robustness, and calibration. These observations have motivated calls for broader evaluation protocols and cross-dataset benchmarks~\cite{faber2024mnist,tuggener2022enough} to capture the limitations of ImageNet-centric optimization.

\textsc{Mobile-Friendly Architectures.} Designing compact networks suitable for mobile and edge devices has been a central focus of recent research. Classical lightweight CNNs—including MobileNet~\cite{howard2019mobilenetv3, howard2017mobilenets, sandler2018mobilenetv2}, ShuffleNet~\cite{zhang2018shufflenet, ma2018shufflenetv2}, and GhostNet~\cite{han2020ghostnet, tang2022ghostnetv2}—achieve strong performance while maintaining low FLOPs and parameter counts. Modern variants such as EfficientNet~\cite{tan2019efficientnet, tan2021efficientnetv2}, ConvNeXt~\cite{zhou2020rethinking}, StarNet~\cite{he2019addressnet}, and ConvMixer~\cite{tuggener2022enough} leverage improved scaling, residual connections, and patch-based convolutions to optimize the trade-off between accuracy and efficiency. Hybrid approaches, including MobileViT~\cite{mehta2021mobilevit}, EfficientFormer~\cite{li2022efficientformer}, and MobileOne~\cite{vasu2023mobileone}, incorporate lightweight attention mechanisms to bring transformer capabilities to resource-constrained devices, demonstrating that careful design can make transformers practical for mobile vision.

\textsc{Neural Architecture Search and Optimization.} Automated NAS techniques have further advanced the efficiency of mobile networks. FBNet~\cite{wu2019fbnet, wan2020fbnetv2} and TinyNet~\cite{zhou2020rethinking} utilize differentiable search strategies to jointly optimize accuracy and hardware efficiency, often outperforming manually designed networks for specific deployment targets. Complementary strategies, including pruning~\cite{han2020ghostnet}, quantization~\cite{liu2024lightweight}, and knowledge distillation~\cite{hao2022cdfkd, hinton2015distilling, zhang2019beyourownteacher}, reduce model size and computation without substantial loss in performance. Innovative architectural primitives such as shift-based operations~\cite{he2019addressnet, wu2017shift} and addition-based AdderNet units~\cite{chen2019addernet} illustrate the diverse approaches researchers employ to minimize FLOPs and memory usage in mobile inference.

\textsc{Cross-Dataset Generalization.} Recent work increasingly emphasizes robust evaluation beyond single benchmarks. Kornblith et al.~\cite{kornblith2019transfer} quantify the transferability of ImageNet-pretrained models across diverse datasets, revealing substantial variability in performance. Fang et al.~\cite{fang2023progressimagenet} and Vishniakov et al.~\cite{vishniakov2023convnet} further show that high ImageNet accuracy does not ensure robustness under domain shifts. Techniques such as self-distillation~\cite{zhang2019beyourownteacher}, multi-level feature sharing~\cite{hao2022cdfkd}, and token pruning in vision transformers~\cite{liang2021evit, rao2021dynamicvit} have been proposed to improve generalization while respecting tight computational budgets.

\begin{sloppypar}
\textsc{Benchmarking and Evaluation Practices.} Standardized evaluation frameworks are crucial for fair comparisons of mobile architectures. Surveys by Wang et al.~\citep{wang2025optimizingedgeai, wang2025empowering} and Lin et al.~\citep{lin2023tinyml} provide comprehensive overviews of strategies for deploying AI on resource-constrained devices, including optimized architectures, data pipelines, and system-level scheduling. Beyond ImageNet, datasets such as CIFAR-10/100~\citep{krizhevsky2009learning}, Stanford Dogs~\citep{khosla2011novel}, HAM10000~\citep{tschandl2018ham10000}, SUN~\citep{xiao2010sun}, and Places~\citep{zhou2017places} are increasingly leveraged to assess cross-domain robustness, highlighting the importance of diverse evaluation protocols that decouple architectural design from training-specific confounders.
\end{sloppypar}

Taken together, these studies motivate a systematic evaluation of mobile-friendly vision models under a unified training protocol, focusing on cross-dataset generalization, robustness, and actionable architectural insights.

\section{Models, Datasets and Benchmark Selection}

We select 11 representative mobile-friendly vision models and 7 diverse datasets to build a comprehensive yet manageable evaluation framework. This setup facilitates analysis of architectural diversity and cross-domain robustness, enabling investigation of whether specific design elements generalize across domains or remain specialized to particular datasets.

To ensure fair and reproducible evaluation of both efficiency and representational capacity for mobile applications, we designed a unified training procedure across all models and datasets. Each model is constrained to approximately $2.5\text{M}$ parameters, except FBNet (which retains its out-of-the-box $3.6\text{M}$ parameters due to the practical difficulty of creating a customized $2.5\text{M}$ variant) and MobileViT, whose transformer-based QKV heads are excluded from the count because of parameter accounting conventions~\cite{mehta2021mobilevit, wu2019fbnet}.

All models are trained from scratch for 100 epochs under a unified framework that standardizes the learning rate schedule, optimizer, loss function, and data pipeline. Specifically, all experiments adopt the Adam optimizer~\cite{menghani2023efficient} with an initial learning rate of $10^{-3}$ (except for MobileViT, which uses $10^{-4}$ due to its higher sensitivity~\cite{mehta2021mobilevit}) and a minimum learning rate of $10^{-5}$. Training begins with a 5-epoch warm-up phase, followed by cosine annealing rate schedule~\cite{loshchilov2017sgdr} to ensure stable convergence~\cite{liu2024lightweight, wang2025empowering}. Input images are resized to $224 \times 224$ and augmented through a consistent pipeline (CutMix~\cite{yun2019cutmix}, RandomFlip, and ColorJitter~\cite{krizhevsky2012imagenet}) to avoid dataset-specific biases. This controlled setup isolates architectural effects from variations in hyperparameters or model capacity. All training is conducted on a desktop equipped with an NVIDIA GeForce RTX 3090 GPU, with a batch size of 32 due to GPU memory constraints.

To rigorously capture these aspects, we introduce the \textbf{Cross-Dataset Score (xScore)}, a metric that quantifies a model’s relative performance across datasets with an explicit penalty for inconsistency. \textbf{xScore} enables systematic identification of architectures that maintain robustness across domains, offering a principled assessment of generalization in mobile-oriented models.

\subsection{Selection of Mobile-Friendly Models}

The eleven selected mobile-friendly vision models are chosen to cover a diverse range of architectures and design philosophies for efficient image classification~\cite{wang2025optimizingedgeai,wang2025empowering,liu2024lightweight,menghani2023efficient}. Classical lightweight CNNs, including \textbf{MobileNet}~\cite{howard2017mobilenets,sandler2018mobilenetv2,howard2019mobilenetv3}, \textbf{ShuffleNet}~\cite{zhang2018shufflenet,ma2018shufflenetv2}, and \textbf{GhostNet}~\cite{han2020ghostnet,tang2022ghostnetv2}, are incorporated for their widespread adoption and efficiency in mobile and embedded scenarios. Modern CNNs, such as \textbf{EfficientNet}~\cite{tan2019efficientnet,tan2021efficientnetv2}, \textbf{ConvNeXt}~\cite{zhou2020rethinking}, and \textbf{StartNet}~\cite{he2019addressnet}, are included to study the effects of improved residual blocks, compound scaling, and training stability on cross-dataset performance. \textbf{ConvMixer}~\cite{tuggener2022enough} offers a compact, patch-based, ultra-lightweight alternative, demonstrating different strategies for reducing computational cost. Neural Architecture Search (NAS) models, such as \textbf{TinyNet}~\cite{zhou2020rethinking} and \textbf{FBNet}~\cite{wu2019fbnet,wan2020fbnetv2}, exemplify automatically optimized designs tailored for mobile hardware. Finally, \textbf{MobileViT}~\cite{mehta2021mobilevit}, as a hybrid and transformer-based model, represent emerging attention-based approaches adapted for efficient mobile inference. Table~\ref{tab:model_stats_compact} summarizes the average parameter counts and FLOPs for all models, which are listed in alphabetical order by model name. Together, these models span a broad range of design choices, enabling systematic evaluation of architectural factors that drive cross-dataset generalization~\cite{fang2023progressimagenet,vishniakov2023convnet,tuggener2022enough}.

\begin{table}[ht]
\centering
\begin{tabular}{lcc}
\toprule
\textbf{Model} & \textbf{Params (M)} & \textbf{FLOPs (M)} \\
\midrule
ConvMixer    & 2.38 & 2347 \\
ConvNext     & 2.65 & 221  \\
EfficientNet & 2.58 & 542  \\
FBNet        & 3.66 & 396  \\
GhostNet     & 2.35 & 314  \\
MobileNet    & 2.31 & 253  \\
MobileOne    & 2.25 & 559  \\
MobileViT*   & 2.82 & 2100 \\
ShuffleNet   & 2.22 & 326  \\
StartNet     & 2.69 & 424  \\
TinyNet      & 2.52 & 253  \\
\bottomrule
\end{tabular}
\caption{Average parameters (M) and FLOPs (M) for mobile models. MobileViT* excludes Transformer QKV heads parameter count.}
\label{tab:model_stats_compact}
\end{table}

\subsection{Selection of Datasets}
To evaluate the cross-dataset generalization of mobile-friendly vision architectures, we employ seven diverse datasets. These datasets span object-centric and scene-centric domains, include coarse- and fine-grained classification, and cover both natural and specialized imagery. Specifically, Imagenette-160 \cite{imagenette} serves as a lightweight proxy for ImageNet-1k, maintaining its object-centric composition and visual diversity while enabling efficient experimentation. CIFAR-10 and CIFAR-100 \cite{krizhevsky2009learning} cover low-resolution general object recognition, whereas Stanford Dogs \cite{khosla2011novel} evaluates fine-grained classification at medium resolution. HAM10k \cite{tschandl2018ham10000} introduces a medical imaging domain to test robustness under strong domain shift, while MIT Indoor-67 \cite{quattoni2009recognizing} and MiniPlaces \cite{zhou2017places} provide complementary scene-level diversity. This dataset suite balances task variety, scale, and domain complexity, providing a balanced foundation for assessing cross-dataset robustness. Table~\ref{tab:datasets} summarizes the datasets and their approximate number of training images per class.

\begin{table}[htbp]
\centering
\resizebox{\textwidth}{!}{%
\begin{tabular}{|l|l|c|c|c|c|}
\hline
\textbf{Dataset} & \textbf{Domain / Type} & \textbf{\# Classes} & \textbf{Train Size} & \textbf{Test Size} & \textbf{\shortstack{Train Images\\per Class}} \\
\hline
\textbf{Imagenette-160} & Natural objects (ImageNet subset) & 10  & 9,469  & 3,925 & $\sim$947 \\
CIFAR-10 & General objects (low-res) & 10  & 50,000 & 10,000 & 5,000 \\
\textbf{CIFAR-100} & General objects (low-res, fine-grained) & 100 & 50,000 & 10,000 & 500 \\
\textbf{Stanford Dogs} & Fine-grained natural objects & 120 & 12,000 & 8,580 & 100 \\
\textbf{HAM10k} & Medical dermoscopic images & 7 & 7,000 & 3,000 & $\sim$1,000 \\
MIT Indoor-67 & Indoor scene classification & 67 & 5,360 & 1,340 & $\sim$80 \\
MiniPlaces & Scene-centric natural images & 100 & 100,000 & 10,000 & 1,000 \\
\hline
\end{tabular}%
}
\caption{Summary of the seven datasets used for cross-dataset generalization evaluation.}
\label{tab:datasets}
\end{table}

\subsection{Cross-Dataset Score as Benchmark}

Evaluating a model on a single dataset can be misleading, as it may perform well on some datasets but poorly on others. To capture both the overall performance and robustness across multiple datasets, we define a cross-dataset performance score, \emph{xScore}, that balances accuracy and consistency. 

Let $M = \{m_1, \dots, m_K\}$ be the set of models and $D = \{d_1, \dots, d_N\}$ the set of datasets. The accuracy of model $m_i$ on dataset $d_j$ is $A_{i,j}$. We define:

\begin{equation}
\hat{A}_{i,j} = \frac{A_{i,j} - \min_i A_{i,j}}{\max_i A_{i,j} - \min_i A_{i,j}}
\label{eq:normalize}
\end{equation}

\begin{equation}
G_i = \frac{1}{N} \sum_{j=1}^{N} \hat{A}_{i,j}, \quad
V_i = \mathrm{Var}\!\left( \{ \hat{A}_{i,j} \}_{j=1}^{N} \right)
\label{eq:meanvar}
\end{equation}

\begin{equation}
xScore_i = G_i - \lambda \cdot V_i, \quad \lambda \in [0,1]
\label{eq:xdg}
\end{equation}

\noindent where:
\begin{itemize}
    \item $\hat{A}_{i,j}$ is the normalized performance value in [0,1] of model $m_i$ on dataset $d_j$ (Eq.~\ref{eq:normalize}),
    \item $G_i$ and $V_i$ measure average performance and consistency of model $m_i$ across datasets (Eq.~\ref{eq:meanvar}),
    \item $xScore_i$ combines accuracy and consistency to quantify cross-dataset robustness of model $m_i$ (Eq.~\ref{eq:xdg}).
\end{itemize}

\section{Experimental Results and Analysis}

\begin{table}[htbp]
\centering
\begin{tabular}{|l|c|c|c|c|c|c|c|}
\hline
\textbf{Model} & \textbf{CIFAR-10} & \textbf{Imagenette-160} & \textbf{CIFAR-100} & \textbf{HAM10k} & \textbf{Dogs} & \textbf{Miniplaces} & \textbf{Indoor-67} \\
\hline
ConvMixer         & 94.52 & \cellcolor{lightgreen}{89.25} & 77.58 & \cellcolor{lightgreen}{80.64} & 54.81 & 58.36 & 57.82 \\ \hline
EfficientNet      & 95.11 & 88.82 & \cellcolor{lightgreen}{78.79} & 79.44 & \cellcolor{lightgreen}{63.78} & \cellcolor{lightgreen}{61.14} & 59.70 \\ \hline
MobileViT         & \cellcolor{lightgreen}{95.19} & 86.55 & 77.25 & 77.38 & 59.18 & 54.78 & \cellcolor{lightgreen}{59.74} \\ \hline
GhostNet          & 93.56 & 86.29 & 73.52 & 76.31 & 55.22 & 57.73 & 41.34 \\ \hline
MobileNet         & 94.17 & 86.24 & 75.74 & 78.78 & 57.53 & 57.39 & 50.60 \\ \hline
TinyNet           & 94.90 & 85.30 & 78.21 & \cellcolor{lightred}{74.38} & \cellcolor{lightred}{29.59} & 59.06 & 50.67 \\ \hline
StartNet          & 94.88 & 84.31 & 77.59 & 79.04 & 48.91 & 55.45 & 51.79 \\ \hline
ShuffleNet        & 92.15 & 84.00 & 75.80 & 78.64 & 55.49 & 57.03 & 53.81 \\ \hline
FBNet             & 94.36 & 83.11 & 75.62 & 78.44 & 51.41 & 57.61 & 48.73 \\ \hline
MobileOne         & 93.34 & 81.40 & 74.55 & 79.11 & 46.74 & 56.33 & 43.51 \\ \hline
ConvNext          & \cellcolor{lightred}{91.97} & \cellcolor{lightred}{76.05} & \cellcolor{lightred}{68.91} & 78.58 & 30.96 & \cellcolor{lightred}{51.01} & \cellcolor{lightred}{32.76} \\ \hline
\end{tabular}
\caption{Cross-dataset accuracy (\%) of 11 mobile-scale models across seven benchmarks.}
\label{tab:xdg_results}
\end{table}

\begin{table}[h!]
\centering
\begin{tabular}{lcccc}
\hline
\textbf{Model} & \textbf{G} & \textbf{V} & \textbf{7-dataset xScore} & \textbf{4-dataset xScore} \\
\hline
\textbf{EfficientNet\_2m} & \textbf{0.964} & \textbf{0.005} & \textbf{0.962} & \textbf{0.925} \\
\textbf{ConvMixer-512-8}  & \textbf{0.861} & \textbf{0.017} & \textbf{0.853} & \textbf{0.876} \\
MobileVIT        & 0.809 & 0.025 & 0.796 & 0.727 \\
MobileNet        & 0.708 & 0.004 & 0.706 & 0.696 \\
StartNet         & 0.695 & 0.028 & 0.681 & 0.692 \\
FBNet            & 0.641 & 0.004 & 0.639 & 0.641 \\
ShuffleNet       & 0.596 & 0.062 & 0.565 & 0.637 \\
GhostNet         & 0.539 & 0.038 & 0.521 & 0.520 \\
MobileOne        & 0.512 & 0.016 & 0.504 & 0.501 \\
TinyNet          & 0.573 & 0.163 & 0.491 & 0.492 \\
ConvNext         & 0.102 & 0.063 & 0.070 & 0.066 \\
\hline
\end{tabular}
\caption{Cross-Dataset Scores (xScore) of Models. The xScores are computed using $\lambda = 0.5$ to balance mean performance and variance across datasets. The last column shows the predicted xScore using the best 4 representative datasets.}
\label{tab:xscores}
\end{table}

\subsection{Do Mobile Architectures Generalize, or Just Specialize?}

Evaluating lightweight models on a single benchmark often overlooks domain-specific biases and fails to capture real-world robustness, providing limited assurance of practical utility across diverse deployment scenarios. To address this limitation, the Cross-Dataset Score ($xScore$) was introduced as a unified metric that jointly measures relative accuracy and consistency across datasets. The normalization step (Eq.~\ref{eq:normalize}) removes scale disparities among datasets, enabling fair comparison. The mean term $G_i$ (Eq.~\ref{eq:meanvar}) reflects a model’s average competitive performance, while the variance term $V_i$ penalizes instability, capturing sensitivity to domain shifts. Combining these components as $xScore_i = G_i - \lambda V_i$ (Eq.~\ref{eq:xdg}) balances accuracy and robustness according to the bias–variance principle, rewarding models that achieve both strong performance and reliable generalization across heterogeneous datasets.

Table~\ref{tab:xdg_results} summarizes the cross-dataset accuracies of eleven mobile-scale models evaluated across seven benchmarks, arranged roughly by dataset popularity and difficulty. For clarity, each dataset’s highest and lowest accuracies are highlighted in green and red, respectively. These column-wise extrema are also used in the $xScore$ calculation and can serve as fixed anchors for evaluating future models, ensuring consistent comparisons without altering the relative scores of the existing set. Table~\ref{tab:xscores} reports each model’s $xScore$, sorted by value. Despite identical training pipelines and comparable parameter budgets, the models exhibit substantial divergence in $xScore$, highlighting pronounced differences in cross-domain generalization.

EfficientNet emerges as the most robust and consistent performer, attaining an $xScore$ of 0.962, followed by ConvMixer (0.853), together forming the top-performing tier. MobileViT, MobileNet, StartNet, and FBNet constitute a second tier of moderately generalizable architectures, with $xScores$ between 0.65 and 0.8. ShuffleNet, GhostNet, MobileOne, and TinyNet form a lower tier, showing weaker generalization with $xScores$ from 0.5 to 0.6. ConvNeXt consistently records the lowest $xScores$, reflecting limited cross-domain robustness.

These discrepancies underscore the variability in generalization among lightweight architectures. Top performers\allowbreak---EfficientNet and ConvMixer\allowbreak---maintain competitive accuracy across both object- and scene-centric datasets, whereas others show pronounced bias toward their training distributions. Some of such bias likely arises from dataset-specific inductive priors or NAS-driven specialization (e.g., FBNet, TinyNet). Notably, MobileViT does not outperform its CNN-based counterparts, despite incorporating a multi-head transformer at each layer to capture global dependencies atop CNN-captured local features\allowbreak---whose parameters are excluded from the total count.

It is important to emphasize that $xScore$ is meaningful only when comparing models of similar parameter count. A lower-capacity model with more parameters may achieve higher raw accuracy on certain datasets than a higher-capacity model with fewer parameters, which could lead to misleading conclusions.

To extend $xScore$ as a practical benchmark for future models of comparable capacity, we selected four datasets from the original seven through a brute-force comparison across all eleven models, identifying the subset that best reproduces the full $xScore$ rankings. The resulting four datasets—\textbf{Imagenette-160}, \textbf{CIFAR-100}, \textbf{HAM10k}, and \textbf{Stanford-Dogs}—achieve the closest match to the full seven-dataset rankings (as shown in Table~\ref{tab:xscores}) and thus serve as a lightweight yet reliable proxy for cross-domain evaluation. Notably, each dataset originates from a distinct visual domain, despite their selection being determined purely by quantitative criteria. For evaluating future mobile models of similar capacity, $xScore$ can be computed using Eq.~\ref{eq:normalize}–Eq.~\ref{eq:xdg}, with normalization bounds $\min(A_{i,j})$ and $\max(A_{i,j})$ fixed to the reference values in Table~\ref{tab:xdg_results}.

\subsection{Is ImageNet a Fair Benchmark for Mobile Models?}

\begin{figure}[htbp]
  \centering
  \includegraphics[width=0.65\linewidth]{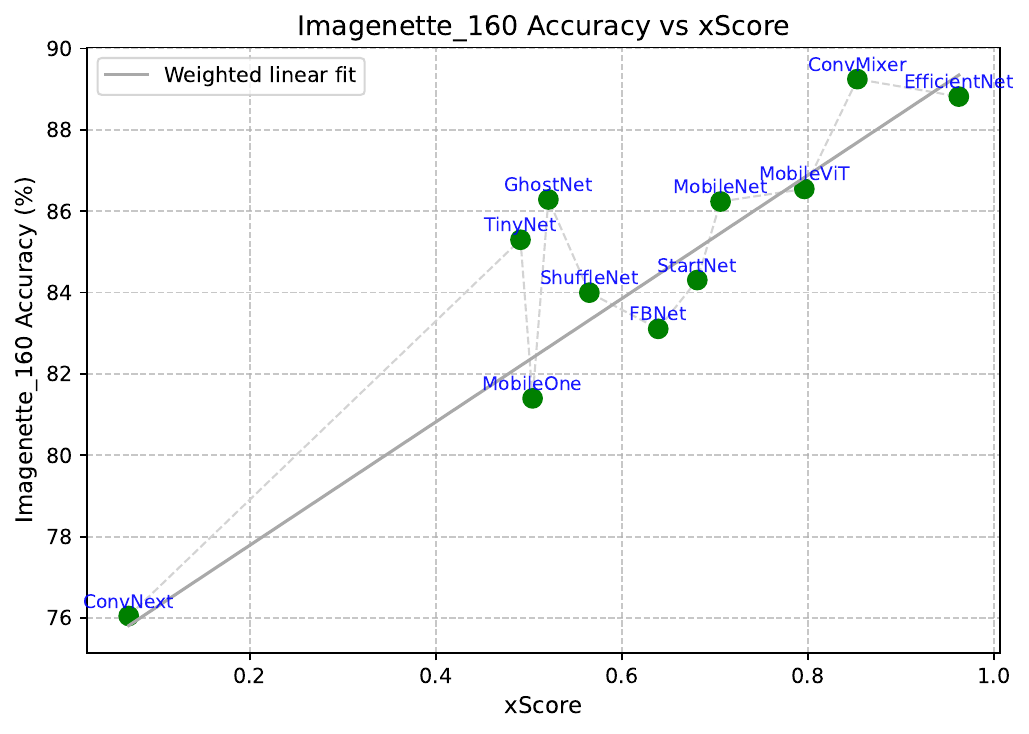}
  \caption{Correlation between xScore and Imagenette-160 accuracy across models.}
  \label{fig:xscore_corr}
\end{figure}

The results in Tables~\ref{tab:xdg_results} and ~\ref{tab:xscores} naturally raise the question of whether ImageNet is a fair benchmark for mobile model performance and robustness. Figure~\ref{fig:xscore_corr} illustrates the relationship between each model’s Imagenette accuracy and its corresponding $xScore$, with $xScore$ values sorted in ascending order along the x-axis. The blue regression line shows a generally positive correlation, suggesting that higher Imagenette accuracy often aligns with stronger cross-domain generalization. However, notable deviations—particularly for TinyNet and GhostNet—demonstrate that strong single-dataset performance does not necessarily indicate robustness across diverse domains.

If ImageNet were a reliable proxy for generalization, models excelling on Imagenette—a compact, structurally similar subset of ImageNet-1k—would also achieve high $xScore$ values. The observed divergence challenges this assumption, especially for mobile-scale architectures with limited parameter budgets, where constrained capacity can limit generalization despite strong single-benchmark performance.

These observations reinforce the importance of evaluating models with metrics like $xScore$, which integrate both performance and consistency across multiple datasets. Furthermore, $xScore$ comparisons are most meaningful among models of similar parameter count, ensuring that robustness, rather than raw parameter-driven accuracy, is being measured. In this light, ImageNet alone provides an incomplete picture of real-world generalization, whereas cross-dataset evaluation offers a more faithful assessment of mobile architectures’ practical robustness.

Building on these insights, evaluating a future mobile model’s $xScore$ using the four representative datasets\allowbreak—--\textbf{Imagenette-160}, \textbf{CIFAR-100}, \textbf{HAM10k}, and \textbf{Stanford-Dogs}\allowbreak—--rather than ImageNet provides a fairer and more efficient assessment protocol. This combined benchmark comprises approximately 79,000 training samples across four diverse domains with 327 classes, compared to ImageNet’s 1.28 million samples spanning 1,000 classes. The reduced yet diverse setup better matches the learning capacity of mobile models, avoiding the underfitting that occurs when parameter limited architectures are trained on full ImageNet.  

\subsection{What Defines Robustness in Mobile Vision Architectures? }

\begin{table*}[ht]
\centering
\scriptsize
\renewcommand{\arraystretch}{1.1}
\setlength{\tabcolsep}{3pt}
\resizebox{\textwidth}{!}{%
\begin{tabular}{>{\raggedright\arraybackslash}p{2.4cm}ccccccccccc}
\toprule
\textbf{Architecture Elements} & \textbf{EffNet} & \textbf{ConvMixer} & \textbf{MobileViT} & \textbf{MobileNet} & \textbf{ShuffleNet} & \textbf{GhostNet} & \textbf{TinyNet} & \textbf{MobileOne} & \textbf{FBNet} & \textbf{ConvNeXt} & \textbf{StarNet} \\
\midrule
Patch-based Stem              & $\times$ & \checkmark & $\times$ & $\times$ & $\times$ & $\times$ & $\times$ & $\times$ & $\times$ & \checkmark & $\times$ \\
Conv Stem                     & \checkmark & $\times$ & \checkmark & \checkmark & \checkmark & \checkmark & \checkmark & \checkmark & \checkmark & $\times$ & \checkmark \\
Depthwise Conv                & \checkmark & \checkmark & \checkmark & \checkmark & \checkmark & \checkmark & \checkmark & \checkmark & \checkmark & \checkmark & \checkmark \\
Pointwise Conv (1×1)          & \checkmark & \checkmark & \checkmark & \checkmark & \checkmark & \checkmark & \checkmark & \checkmark & \checkmark & \checkmark & \checkmark \\
Width/Channel Scaling         & \checkmark & $\times$ & $\times$ & \checkmark & \checkmark & \checkmark & \checkmark & \checkmark & $\times$ & $\times$ & \checkmark \\
Residual Skip                 & \checkmark & \checkmark & \checkmark & \checkmark & $\times$ & \checkmark & \checkmark & \checkmark & \checkmark & \checkmark & \checkmark \\
Inverted Residual             & \checkmark & $\times$ & \checkmark & \checkmark & $\times$ & \checkmark & \checkmark & $\times$ & \checkmark & $\times$ & $\times$ \\
C--H--W Scaling               & \checkmark & $\times$ & $\times$ & $\times$ & $\times$ & $\times$ & $\times$ & $\times$ & $\times$ & $\times$ & $\times$ \\
Squeeze-and-Excite (SE)       & \checkmark & $\times$ & $\times$ & $\times$ & $\times$ & \checkmark & \checkmark & \checkmark & $\times$ & $\times$ & $\times$ \\
Channel-wise Concatenation    & $\times$ & $\times$ & $\times$ & $\times$ & \checkmark & \checkmark & $\times$ & $\times$ & $\times$ & $\times$ & $\times$ \\
NAS                           & \checkmark & $\times$ & $\times$ & $\times$ & $\times$ & $\times$ & \checkmark & $\times$ & \checkmark & $\times$ & $\times$ \\
ReLU+                         & \checkmark & \checkmark & \checkmark & \checkmark & $\times$ & \checkmark & \checkmark & $\times$ & \checkmark & \checkmark & \checkmark \\
Layerwise Multi-head Transformer & $\times$ & $\times$ & \checkmark & $\times$ & $\times$ & $\times$ & $\times$ & $\times$ & $\times$ & $\times$ & $\times$ \\
Global Avg Pooling            & \checkmark & \checkmark & \checkmark & \checkmark & \checkmark & \checkmark & \checkmark & \checkmark & \checkmark & \checkmark & \checkmark \\
FCN Classifier                & \checkmark & \checkmark & \checkmark & \checkmark & \checkmark & \checkmark & \checkmark & \checkmark & \checkmark & \checkmark & \checkmark \\
\bottomrule
\end{tabular}%
}
\caption{Comparison of core architectural elements across modern lightweight convolutional and hybrid models.}
\label{tab:arch_elements}
\end{table*}

Table~\ref{tab:arch_elements} lists the key architectural elements in the first column and indicates which models incorporate each component in the subsequent columns. All models follow a multi-layer paradigm, progressively distilling feature maps using depthwise and pointwise convolutions, along with residual skips, to capture and condense local patterns. Except for ShuffleNet, all models employ some variation of the standard ReLU activation. In every case, the final-layer feature map, with shape [B, C, H, W], is aggregated via global average pooling to [B, C] before being passed to a fully connected (FCN) classifier.

While these common building blocks effectively save parameters and boost performance across models, EfficientNet and ConvMixer leverage additional architectural design choices that drive their outstanding performance.

EfficientNet distinguishes itself through joint scaling of depth, width, and resolution across layers, progressively increasing channel numbers while reducing spatial dimensions. This approach maintains balanced representational capacity without bottlenecks or wasted parameters. The integration of Squeeze-and-Excite (SE) attention and inverted residual connections further enhances channel-wise information flow as spatial dimensions shrink, ensuring efficient feature extraction throughout the network hierarchy.

ConvMixer, in contrast, follows a fundamentally different design. It partitions the input image into non-overlapping patches, transforming it into an isotropic feature tensor of fixed spatial dimensions. Each subsequent layer updates this tensor through three core operations—depthwise convolution, residual addition, and pointwise convolution. By maintaining constant spatial resolution, ConvMixer efficiently captures local patterns within each channel without the typical spatial information loss caused by downsampling. This isotropic representation enables strong accuracy and robustness despite its remarkably simple and uniform design.

Together, EfficientNet and ConvMixer exemplify complementary strategies for channel-wise information processing. EfficientNet compresses and refines information through hierarchical scaling and SE-driven reweighting, whereas ConvMixer expands tensors directly to the target channel dimension, preserving large spatial maps (16×16 or higher) without SE or inverted residuals. This design allows free channel-wise information flow but incurs higher per-layer parameters and FLOPs. Consequently, on high-resolution datasets such as Stanford Dogs, ConvMixer may need larger patch sizes to reach the target spatial resolution, risking information loss and reduced accuracy. Under comparable parameter budgets, EfficientNet can maintain greater depth, yielding superior representational efficiency and overall performance.

\section{Conclusion and Future work}

In this study, we systematically evaluated eleven modern mobile-friendly vision architectures across seven diverse datasets, emphasizing cross-dataset generalization rather than single-benchmark performance. We find that single-benchmark metrics, such as ImageNet or Imagenette accuracy, do not fully capture cross-dataset robustness, highlighting the limitations of evaluating mobile models on isolated datasets. To address this, we introduce the Cross-Dataset Score ($xScore$) as a unified metric for both accuracy and robustness.

EfficientNet and ConvMixer consistently emerge as top performers, exhibiting strong accuracy and stable $xScore$ across object- and scene-centric datasets. EfficientNet achieves robustness through hierarchical scaling, SE attention, and inverted residual connections, effectively balancing channel and spatial dimensions. ConvMixer, by contrast, employs isotropic patch-based representations with depthwise and pointwise convolutions, providing a simple yet effective approach for channel-wise information flow without hierarchical compression. These complementary strategies demonstrate that architectural design choices, rather than parameter count alone, govern cross-domain generalization in mobile models.

We also propose a practical four-dataset evaluation protocol-\textbf{Imagenette-160}, \textbf{CIFAR-100}, \textbf{HAM10k}, and \textbf{Stanford-Dogs}-which closely approximates the full $xScore$ while substantially reducing computational cost. This lightweight benchmark offers a capacity-aligned framework for assessing mobile architectures and encourages the development of models that generalize reliably across diverse real-world domains.

Currently, $xScore$ was evaluated under a single parameter configuration. Future work should systematically vary model capacity to examine the relationship between $xScore$ and parameter count, revealing whether higher capacity consistently improves cross-dataset generalization or whether certain designs achieve high $xScore$ efficiently at lower budgets. This analysis could also uncover diminishing returns, inform capacity-aware architecture selection, and guide the design of mobile models that balance robustness, accuracy, and efficiency. Moreover, while classification often benefits from channel-wise average pooling that compresses spatial information, tasks such as semantic segmentation and object detection demand architectures that preserve fine-grained spatial details, motivating further research into how $xScore$ varies across models for these spatially intensive tasks.

In summary, cross-dataset evaluation reveals that robustness in mobile vision models is driven by architectural design choices—such as hierarchical scaling, patch-based representations, and channel-wise attention mechanisms—rather than parameter count alone. The $xScore$ framework provides a principled metric to guide future architecture design and selection for practical, mobile-constrained deployment scenarios.

\end{document}